%% file: main.tex
\documentclass[sigconf,authorversion]{acmart}

\AtBeginDocument{%
  \providecommand\BibTeX{{%
    \normalfont B\kern-0.5em{\scshape i\kern-0.25em b}\kern-0.8em\TeX}}}

\copyrightyear{2024}
\acmYear{2024}
\setcopyright{acmlicensed}\acmConference[ETRA '24]{2024 Symposium on Eye Tracking Research and Applications}{June 4--7, 2024}{Glasgow, United Kingdom}
\acmBooktitle{2024 Symposium on Eye Tracking Research and Applications (ETRA '24), June 4--7, 2024, Glasgow, United Kingdom}
\acmDOI{10.1145/3649902.3653518}
\acmISBN{979-8-4007-0607-3/24/06}

\input{own_packages_and_definitions}

\begin{document}
\title{NMF-Based Analysis of Mobile Eye-Tracking Data}

\author{Daniel Klötzl}
\email{Daniel.Kloetzl@visus.uni-stuttgart.de}
\orcid{0000-0002-4222-3320}
\affiliation{%
  \institution{University of Stuttgart}
  \country{Germany}
}
\author{Tim Krake}
\email{Tim.Krake@visus.uni-stuttgart.de}
\orcid{0009-0004-7084-3633}
\affiliation{%
  \institution{University of Stuttgart}
  \country{Germany}
}
\author{Frank Heyen}
\email{Frank.Heyen@visus.uni-stuttgart.de}
\orcid{0000-0002-5090-0133}
\affiliation{%
  \institution{University of Stuttgart}
  \country{Germany}
}
\author{Michael Becher}
\email{Michael.Becher@visus.uni-stuttgart.de}
\orcid{0000-0002-0072-1655}
\affiliation{%
  \institution{University of Stuttgart}
  \country{Germany}
}
\author{Maurice Koch}
\email{Maurice.Koch@visus.uni-stuttgart.de}
\orcid{0000-0003-0469-8971}
\affiliation{%
  \institution{University of Stuttgart}
  \country{Germany}
}
\author{Daniel Weiskopf}
\email{Daniel.Weiskopf@visus.uni-stuttgart.de}
\orcid{0000-0003-1174-1026}
\affiliation{%
  \institution{University of Stuttgart}
  \country{Germany}
}
\author{Kuno Kurzhals}
\email{Kuno.Kurzhals@visus.uni-stuttgart.de}
\orcid{0000-0003-4919-4582}
\affiliation{%
  \institution{University of Stuttgart}
  \country{Germany}
}

\begin{abstract}
The depiction of scanpaths from mobile eye-tracking recordings by thumbnails from the stimulus allows the application of visual computing to detect areas of interest in an unsupervised way.
We suggest using nonnegative matrix factorization (NMF) to identify such areas in stimuli.
For a user-defined integer $k$, NMF produces an explainable decomposition into $k$ components, each consisting of a spatial representation associated with a temporal indicator.
In the context of multiple eye-tracking recordings, this leads to $k$ spatial representations, where the temporal indicator highlights the appearance within recordings.
The choice of $k$ provides an opportunity to control the refinement of the decomposition, i.e., the number of areas to detect.
We combine our NMF-based approach with visualization techniques to enable an exploratory analysis of multiple recordings.
Finally, we demonstrate the usefulness of our approach with mobile eye-tracking data of an art gallery.
\end{abstract}



\maketitle

\begin{figure*}
\includegraphics[width=\textwidth]{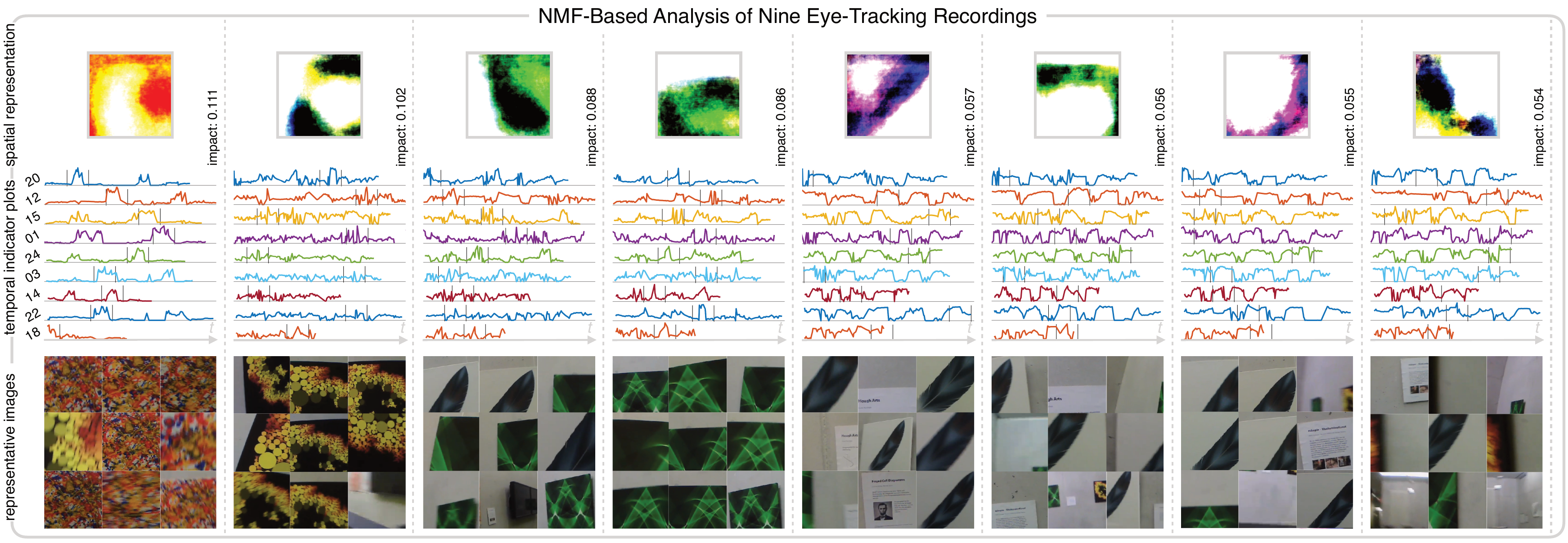}
\caption{Our visual analysis technique applied to nine mobile eye-tracking recordings of a scene from an art gallery.
Based on the underlying nonnegative matrix factorization (NMF), the preprocessed recordings are decomposed into eight spatiotemporal components, where each is described by a spatial representation, temporal indicator plots, and an impact.
The combination of these two representations provides a spatiotemporal clustering of the recordings.
To link the eight clusters to the recordings, a representative image from each recording is assigned to it based on the highest peak of the respective temporal indicator plots.
}
\label{fig:teaser}
\end{figure*}

\section{Introduction}\label{sec:intro}
A common challenge in eye-tracking research is to answer research questions regarding common behavior in recorded gaze data.
Typical examples are classifying differences between novices and experts~\cite{Castner2020} and identifying different visual task solution strategies~\cite{Kumar2019task}.
To achieve this goal, a vast set of different statistical methods~\cite{Holmqvist2011} and visualization/visual analytics~\cite{Blascheck2017} approaches have been developed over the years.
Many of these approaches rely on defining areas of interest (AOIs), which requires additional semantic annotations, often accompanied by tedious manual labeling.
Alternatively, image-based approaches directly combine gaze and visual stimulus for calculations on the recordings without annotations~\cite{Kurzhals2016, Koch2022}. 
In the past, image patches were used to train machine learning models~\cite{SteilHB18}, for clustering~\cite{Castner2020}, and as support for analysis tasks~\cite{Kurzhals2021}. 
Most of these approaches are based on extracted features, either traditional ones, such as histograms and SIFT \cite{Lowe1999}, or feature vectors derived from neural networks, such as ResNet~\cite{He2016}. 
While this direction often provides good results, an interpretation of intermediate steps and general explainability are usually limited.

We propose making use of \textit{nonnegative matrix factorization} (NMF) \cite{Paatero1994} to identify key elements in video recordings of mobile eye tracking and their temporal occurrence in different recordings. 
In particular, our main goal is to summarize AOIs and when they were investigated in different recordings. 
NMF is an approach for multivariate analysis for image processing and clustering~\cite{Hong2016, Yang2017, Yang2020}, computer vision~\cite{Zhang2015}, and audio processing~\cite{Gemmeke2013}.  
The advantage of this technique is that, if applied to image data, the resulting components can be directly interpreted by a human.
Combined with interactive visualization, NMF can serve as an initial analysis step to find AOIs and characterize scanpaths from different recordings.

This work concentrates on an NMF-based approach for the visual analysis of multiple mobile eye-tracking recordings. 
Previous techniques created compact spatial representations of multiple recordings based on scanpath clustering~\cite{Kumar2019clustered} or dimensionality reduction on thumbnail images~\cite{Kurzhals2021}. 
However, these techniques neglected the time aspect. 
Other approaches like Gaze Stripes~\cite{Kurzhals2016} or Gaze Spirals~\cite{Koch2022} retain the temporal component but have limited scalability concerning the number of recordings or participants.

We provide a solution to both problems by generating a compact spatiotemporal representation of multiple recordings. 
While NMF has been used for video analysis before~\cite{Pnevmatikakis2016, Kondo2017}, we propose the application to eye-tracking data by incorporating gaze and fixation information into the decomposition and analysis.
The resulting NMF-based analysis shown in \autoref{fig:teaser} illustrates the decomposition of nine mobile eye-tracking recordings that we took of a scene from an art gallery. 
Using NMF, the dataset is decomposed into eight spatiotemporal components that are visually described by a spatial representation (top row) and indicator plots for each recording (middle row). To visually monitor the indicated timestamps, representative images (bottom row) from each recording are selected based on the highest peak of the respective temporal indicator plots.
In this configuration, our method successfully identifies four of the five AOIs.
We showcase our approach with a dataset recorded for this work. This dataset is publicly available \cite{darus-4023_2024}.

\section{Background}\label{sec:background}
Matrix factorizations (or decompositions) are frequently used, as they provide a way to decompose complex objects into simpler ones~\cite{Pnevmatikakis2016, Kondo2017}.
The most common techniques include factorizations based on the concept of eigenvalues and factorizations related to solving algebraic equations~\cite{Lyche:2020:NumericalLinearAlgebraAndMatrixFactorization}.
For nonnegative data, such as video data with grayscale or RGB values, classical matrix factorizations do not exploit the nonnegative nature of the data.
They may produce components with non-interpretable characteristics or even misleading visual features.
NMF~\cite{Paatero1994,Lee1999} can be used to include nonnegativity~into~the~factorization.

Given a matrix $X \in \mathbb{R}^{n \times m}$ (with not necessarily nonnegative entries), NMF aims to find nonnegative matrices $W \in \mathbb{R}^{n \times k}$ and $H \in \mathbb{R}^{k \times m}$, i.e., $W \geq 0$ and $H \geq 0$, such that $X = W H$.
Due to the parameter $0 < k < \min\{n,m\}$ (defining the low-rank property) and the restriction of the matrices $W$ and $H$ to be nonnegative, the equation is only satisfied in an approximate manner.
Numerically, NMF can be computed via mathematical optimization techniques~\cite{Berry2007} that solve the constrained~minimization~problem $\min_{W,H} \frac{1}{2}\lVert X-WH \rVert^2_F$ subject to $W \geq 0, H \geq 0.$

For our NMF-based approach, we use an implementation based on the MATLAB function \texttt{nnmf}.\footnote{\url{http:/mathworks.com/help/stats/nnmf.html}}
This function follows prior work, see~\cite{Berry2007}, implementing an iterative solver for the constrained~minimization~problem.

\begin{figure*}[t]
\includegraphics[width=\textwidth]{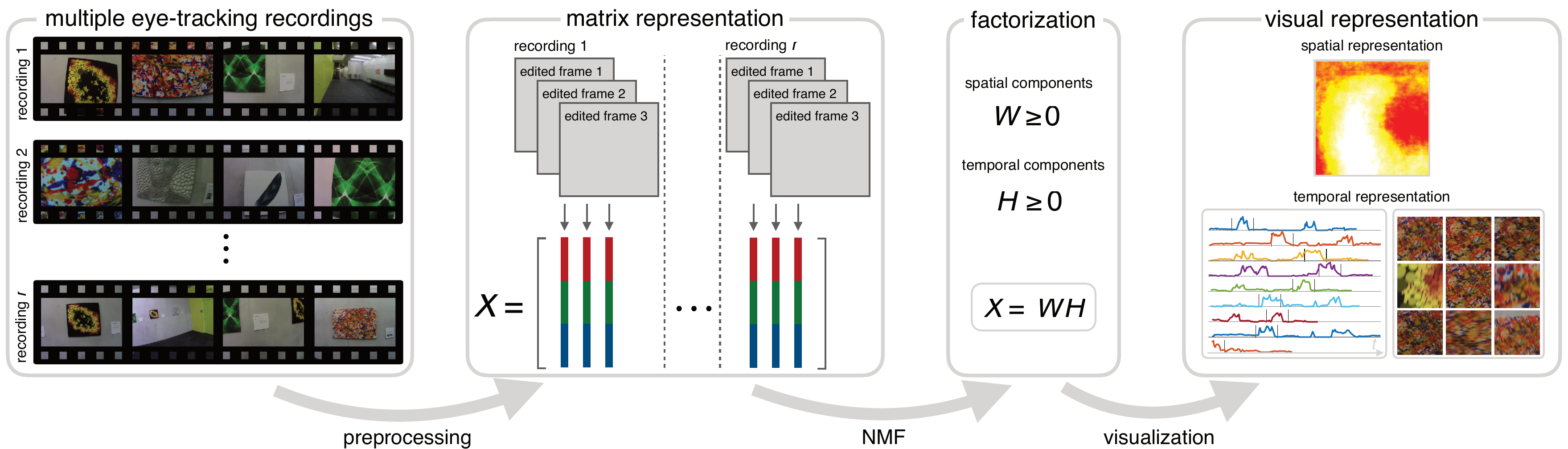}
\caption{Overview of our visually guided approach to identifying spatiotemporal features among multiple eye-tracking recordings:
After preprocessing the recordings (using image patches, focusing on fixations, and vectorizing them into a matrix representation), the edited frames are represented in an overarching matrix that can be factorized via NMF to generate appropriate clusters.
Finally, a clustered representation with interactive visualizations is established to provide an exploratory analysis of the eye-tracking recordings.\vspace{-2ex}
}
\label{fig:overview_approach}
\end{figure*}

\section{Technique}\label{sec:technique}
This section presents our visually guided approach to identifying spatiotemporal features among multiple mobile eye-tracking recordings.
The NMF-based approach provides a clustered overview with visual representations.
\autoref{fig:overview_approach} illustrates the entire process, and the following subsections describe the corresponding steps in more detail.
These steps include the preprocessing for NMF, the application and interpretation of NMF, and the visualization~strategies.

\subsection{Preprocessing for NMF}\label{subsec:Prep_NMF}
Having a description of the eye-tracking recordings, the next step is to prepare the frames for the application of NMF.
This step includes two procedures: the use of image patches and fixations as well as the vectorization into a matrix representation.
There are $r$ recordings with $m_1,\dots,m_r$ number of frames (i.e., with different durations) and a resolution of $n_1 \times n_2$ (width $\times$ height). 
In general, the presented approach can deal with any color model (e.g., grayscale, RGB, or CIELAB values).
Nevertheless, our analysis focuses on RGB~values.

\paragraph{Image Patches and Fixations}
To foster the clustering properties of NMF, it is necessary to adjust the raw video frames by integrating the gaze behavior.
First, we recommend using image patches by cropping a rectangular region from each video frame around the gaze point.
This step establishes a link between the point of interest represented by the gaze and the visual stimulus.
It also supports suppressing irrelevant information, which reduces the risk of producing misleading patterns with NMF.
The result of this procedure is a reduction of the resolution by cropping the video frames, i.e., the new resolutions of the $r$ recordings are given by the following modification:
\begin{displaymath}
n_1 \times n_2
\qquad
\xrightarrow{\text{~~cropping~~}}
\qquad
\tilde{n}_1 \times \tilde{n}_2.
\end{displaymath}

Our second recommendation is to thin out the frames by using fixations. 
Since the fixations aggregate the gaze behavior of a participant, redundant information is filtered out, and important temporal states are emphasized.
The effect of this filtering procedure is a reduction of frames, i.e., the number of frames of the $r$ recordings is modified as follows: \begin{displaymath}
m_1 , \dots , m_r
\qquad
\xrightarrow{\text{~~filtering~~}}
\qquad
\tilde{m}_1 , \dots , \tilde{m}_r.
\end{displaymath}
With the two preprocessing steps, the participant's gaze behavior is naturally integrated into the NMF-based visual analysis pipeline.
This helps identify proper spatiotemporal patterns with NMF.

\paragraph{Vectorization}
Based on the previous steps, we deal with $r$ recordings that have $\tilde{m}_1 , \dots , \tilde{m}_r$ number of frames and a resolution of $\tilde{n}_1 \times \tilde{n}_2$.
To apply NMF, it is necessary to represent the data with matrices.
We aim to combine the preprocessed video frames into a single matrix representation.

As illustrated in \autoref{fig:overview_approach}, for each edited video frame, the pixel values of the color channels (here, RGB values with the three channels R, G, and B) are stacked upon each other, and the resulting vectors of the $r$ recordings are column-wise filled into the universal matrix $X\in \mathbb{R}^{(3\cdot\tilde{n}_1\cdot\tilde{n}_2) \times (\tilde{m}_1+\dots+\tilde{m}_r)}$, i.e.,
\begin{displaymath}
X = \begin{bmatrix}
|           &           & |            &           & |         &           &  |
\\
x^{(1)}_1   & \cdots    & x^{(1)}_{\tilde{m}_1}  & \cdots & x^{(r)}_1    & \cdots    & x^{(r)}_{\tilde{m}_r}
\\
|           &           & |            &           & |         &           &  |
\end{bmatrix}.
\end{displaymath}
This universal representation of the recordings has the following three advantages:
\begin{itemize}[noitemsep,topsep=2pt,leftmargin=*]
\item the problem of having multiple recordings with different durations is bypassed,
\item spatiotemporal clusters among the recordings are identified simultaneously, and
\item the matrix structure facilitates explainability and interpretability of upcoming results.
\end{itemize}
The process of vectorization concludes the preprocessing of the eye-tracking recordings. 
The resulting universal matrix $X$ combines recorded video frames and relevant gaze data.

\begin{figure*}[t]
\includegraphics[width=\textwidth]{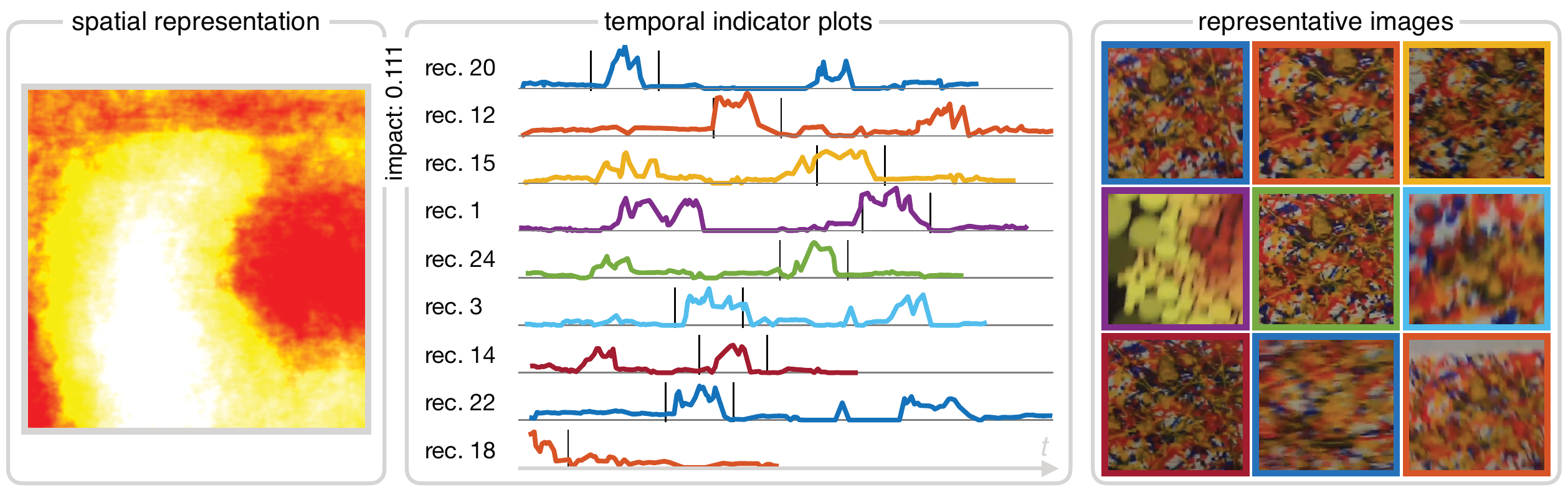}
\caption{
Overview of our strategy to visualize components in our NMF-based visual analysis of eye-tracking recordings. 
The extracted spatiotemporal patterns belong to the most dominant component in \autoref{fig:teaser}, where nine eye-tracking recordings are factorized into \emph{k} = 8 components.
The represented component consists of an impact value, a spatial representation, and associated temporal indicator plots. 
Furthermore, representative reference images from the recordings based on the highest peaks (enclosed by two gray lines in the temporal indicator plots) are shown.
The reference images for each recording are displayed side by side and row by row, as the color of the image frames and temporal indicator plots demonstrate.
}
\label{fig:vis_analysis_compts}
\end{figure*}

\subsection{Application and Interpretation of NMF}\label{subsec:App_NMF}
The main result of the preprocessing steps from the previous subsection is the matrix $X \in \mathbb{R}^{(3\cdot\tilde{n}_1\cdot\tilde{n}_2) \times (\tilde{m}_1+\dots+\tilde{m}_r)}$ that consists of the edited and vectorized video frames.
This matrix representation can be processed with NMF.
According to \autoref{sec:background}, the resulting factorization into the nonnegative matrices $W$ and $H$ is given by
{\small
\begin{align*}
&X = W H
=
\begin{bmatrix}
|   &       & |
\\
w_1 & \cdots & w_k
\\
|   &       & |
\end{bmatrix}
\begin{bmatrix}
&\rule[2pt]{12pt}{0.4pt} & h_1^T & \rule[2pt]{12pt}{0.4pt}&
\\
&& \vdots &&
\\
&\rule[2pt]{12pt}{0.4pt} & h_k^T & \rule[2pt]{12pt}{0.4pt}&
\end{bmatrix}
\\
&= 
\begin{bmatrix}
|   &       & |
\\
w_1 & \cdots & w_k
\\
|   &       & |
\end{bmatrix}
\begin{bmatrix}
h^{(1)}_{1,1}   & \cdots    & h^{(1)}_{1,\tilde{m}_1} & \cdots    & h^{(r)}_{1,1}     & \cdots    & h^{(r)}_{1,\tilde{m}_r}
\\
\vdots      &           & \vdots        &           & \vdots        &           & \vdots
\\
h^{(1)}_{k,1}   & \cdots    & h^{(1)}_{k,\tilde{m}_1} & \cdots & h^{(r)}_{k,1}    & \cdots    & h^{(r)}_{k,\tilde{m}_r}
\end{bmatrix},
\end{align*}
}
\hspace{-0.5ex}with spatial components $w_1,\dots,w_k \in \mathbb{R}^{(3\cdot\tilde{n}_1\cdot\tilde{n}_2)}$, temporal components $h_1,\dots,h_k \in \mathbb{R}^{(\tilde{m}_1+\dots+\tilde{m}_r)}$, and user-controllable parameter $k$ (characterizing the low-rank property of the factorization).
The interpretation of the component pairs~$(w_j,h_j)$ and parameter $k$ in the context of multiple eye-tracking recordings is as follows:
\paragraph{Spatial Components} The vectors $w_1,\dots,w_k$ describe the recordings spatially (see dimension $3 \cdot \tilde{n}_1 \cdot \tilde{n}_2$). 
According to the preprocessing, each vector $w_j$ semantically consists of stacked RGB values and, thus, represents a static image.
We refer to this image as the spatial representation of the component $j$, where $1 \leq j \leq k$.  
The resulting $k$ spatial representations (given by $w_1,\dots,w_k$) provide a visual clustering among the eye-tracking recordings.
\paragraph{Temporal Components} The vectors $h_1,\dots,h_k$ describe the recording temporally (see dimension $\tilde{m}_1+\dots+\tilde{m}_r$).
Since the matrix $X$ is filled with the recordings column-wise, each vector $h_j$ covers the aggregated temporal evolution of a spatial representation.
To be more precise, given the spatial representation $w_j$ with $1 \leq j \leq k$, the associated vector $h_j$ should be split into its $\tilde{m}_1,\dots,\tilde{m}_r$ parts to highlight the respective influence, i.e., $\big(h_{j,1}^{(1)},\dots,h_{j,\tilde{m}_1}^{(1)}\big)$, $\big(h_{j,1}^{(2)},\dots,h_{j,\tilde{m}_2}^{(2)}\big)$, $\dots$, and $\big(h_{j,1}^{(r)},\dots,h_{j,\tilde{m}_r}^{(r)}\big)$. 
With these temporal indicators, the appearance of the associated spatial representation can be precisely determined within the recordings. 
\paragraph{Number of Components} The user-defined parameter~$k$ provides an opportunity to control the refinement of the decomposition.
Depending on the complexity of the video recordings, it makes sense to adapt the number of components to the most requested dominant features.

\subsection{Visualization}\label{subsec:Visualization}
Based on the NMF-processed data, a visually guided analysis is performed.
This approach is exemplified by the overview in \autoref{fig:teaser}, where $r=9$ mobile eye-tracking recordings are used, and NMF is applied to them with parameter $k=8$.
This leads to spatial components $w_1,\dots,w_8$ and temporal components $h_1,\dots,h_8$.
We propose using these components for the visual analysis, as described in the following three paragraphs.

\paragraph{Impact of Components}
The first step of the visual analysis is to sort the component pairs $(w_1,h_1),\dots,(w_k,h_k)$ according to their impact.
While there are many different ways to measure the impact, we suggest using the $p$-norm $\lVert h_j \rVert_p$ of the corresponding temporal components $h_1,\dots,h_k$.
Since we want to find component pairs that exhibit both general impact and anomalies, the Euclidean distance with $p=2$ is appropriate.
It considers general impact and is sensitive to outliers.
Having the impact of the component pairs, we can sort them by their influence (\autoref{fig:teaser}).

\paragraph{Spatial and Temporal Representation of Components}
Next, we visualize the component pairs $(w_1,h_1),\dots,(w_k,h_k)$.
A zoomed-in illustration of the visualization strategy is demonstrated with the most dominant component pair $(w_1,h_1)$ in \autoref{fig:vis_analysis_compts} (compare \autoref{fig:teaser}).
We use this strategy for all components $j=1,\dots,k$, as explained in the following.

First, the spatial component $w_j$ is directly presented using the spatial arrangement of the component, i.e., by transforming $w_j$ to its initial resolution $\tilde{n}_1\times \tilde{n}_2$ (\autoref{fig:vis_analysis_compts}, left).
As described in \autoref{subsec:App_NMF}, these images can be understood as a weighted additive blending of image patches from the original frames of the eye-tracking recordings.

Each spatial representation with its underlying spatial component~$w_j$ has an associated temporal component~$h_j$.
According to \autoref{subsec:App_NMF}, a temporal component~$h_j$ stores information about all recordings.
Thus, partitioning these into values $\big(h_{j,1}^{(1)},\dots,h_{j,\tilde{m}_1}^{(1)}\big)$, $\big(h_{j,1}^{(2)},\dots,h_{j,\tilde{m}_2}^{(2)}\big)$, $\dots$, and $\big(h_{j,1}^{(r)},\dots,h_{j,\tilde{m}_r}^{(r)}\big)$, enables the visualization via separate line charts stacked upon each other (\autoref{fig:vis_analysis_compts}, center).
We call these representations ``temporal indicator plots.''
They highlight the spatial representation's appearance within the recordings, which allows users to identify important time ranges.
For better comparability, all temporal indicator plots are normalized individually.
This allows comparing characteristic features across recordings, e.g., their lengths or individual shapes of the temporal components.
The normalization procedure is justified because the impact of component pairs is compared previously.

\paragraph{Linking to Eye-Tracking Recordings}
The spatial representations with associated temporal indicator plots provide a clustering of the recordings. 
While this clustering often reveals clear temporal patterns (via the temporal indicator plots), the related spatial representations are usually hard to interpret.
Hence, it is useful to additionally link component pairs $(w_j,h_j)$ to actual scenes from the eye-tracking recordings.
We propose selecting additional reference images from the respective recordings based on the temporal indicator plots.
More precisely, given a cluster with component pairs $(w_j,h_j)$, each temporal indicator plot is examined with regard to peaks.
While there are sophisticated outlier detection algorithms, we suggest using the highest peak of each indicator plot.
As illustrated in \autoref{fig:vis_analysis_compts}, the highest peak is enclosed by two vertical bars (only one bar if located near the boundaries), and the respective reference frame is extracted and visualized in a matrix view (\autoref{fig:vis_analysis_compts}, right), i.e., the reference images are displayed side by side and row by row (as the color of the image frames and plots demonstrate).

The current implementation is static, but it is planned to integrate interaction into the visualizations.
More precisely, users shall be allowed to interactively choose different positions within the temporal indicator plots to display other reference images.
This interactive exploration will likely improve our analysis approach and might reveal additional features.

\section{Showcase}\label{sec:showcase}
We recorded a dataset of people walking through a small art gallery\footnote{The used eye-tracking recordings are publicly available~\cite{darus-4023_2024}.} to show the usefulness of our approach.
The scenario is depicted in \autoref{fig:artwork_gallery}. 
It comprises multiple AOIs, and related recordings cover multiple scanpath characteristics.

\subsection{Dataset}\label{subsec:dataset}

\begin{figure}[t]
\centering
\includegraphics[width=\columnwidth]{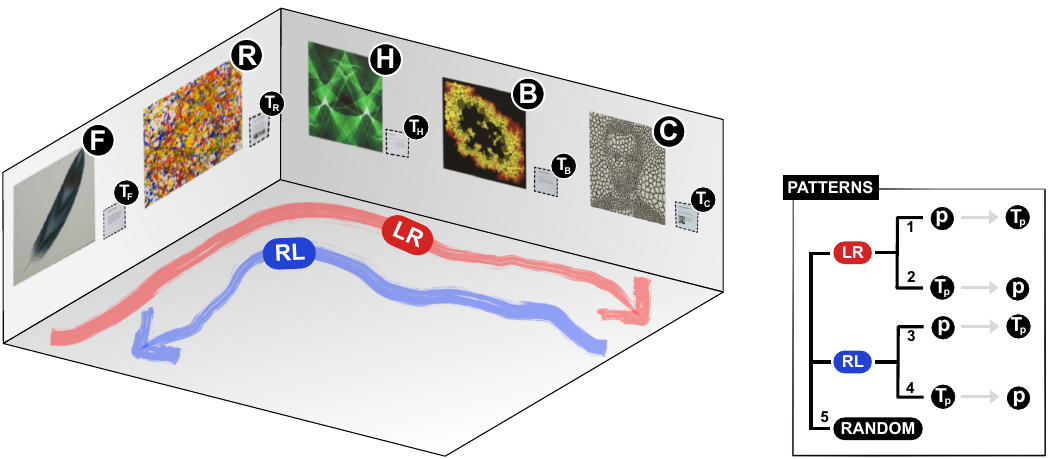}
\caption{Scene of Artwork Gallery with five paintings: \emph{Software Feathers (F)}, \emph{roboPix (R)}, \emph{Hough Arts (H)}, \emph{Bubble Hierarchies (B)}, and \emph{Frayed Cell Diagram (C)}. 
Each painting is accompanied by a plate with text description \emph{Text} (\emph{T}$_\text{\emph{p}}$) with \emph{p} $\mathbf{\in \{}$\emph{F, R, H, B, C}$\mathbf{\}}$. 
The two walking directions \emph{LR} and \emph{RL} define the attendance order. 
The order between a painting and its text description induces a subpattern.}
\label{fig:artwork_gallery}
\end{figure}

\begin{figure*}[tb]
\centering
\includegraphics[width=\textwidth]{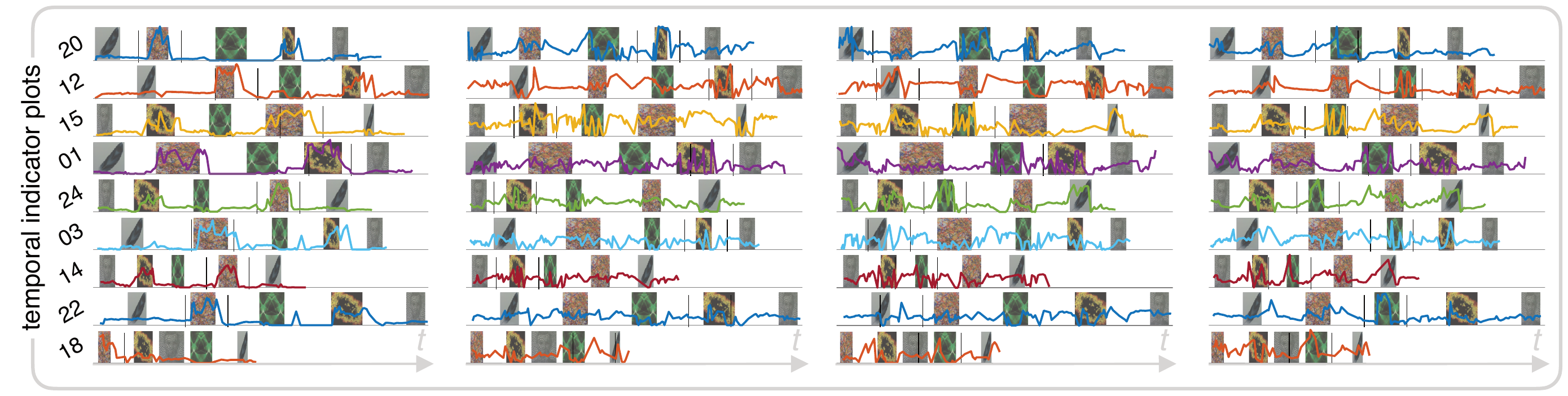}
\caption{Visual comparison of manually labeled AOIs in our NMF-based analysis of nine recordings of a scene from an art gallery. Manual labels are illustrated as small sample images of the five AOIs in the first four temporal indicator plots. This allows evaluating the NMF-based decomposition via their indicator plots. The remaining components, namely the spatial representation and the representative images, are depicted in \autoref{fig:teaser}.}
\label{fig:analysis_best9}
\end{figure*}

Similar recordings were presented in other publications~\cite{Koch2022, Oeney2023, Pathmanathan2023} but are not publicly available due to privacy restrictions in previous recordings. 
The new dataset consists of 27 recordings from three participants, ranging in duration from 50 to 205 seconds. Each participant was recorded nine times, while inspecting the images in predefined orders to create a set of distinctive scanpaths.  
We recorded the data using the Pupil~Invisible~\cite{Tonsen2020} mobile eye-tracking glasses.
For each recording, the dataset provides video from the participant's point of view, the head rotation and acceleration data from the built-in 6-DoF inertial measurement unit (IMU), and the eye position in degrees and pixel coordinates (in reference to the video).

All recordings were trimmed to show the video only from the beginning to the end of the task.
The tasks were designed to produce distinct groups of scanpaths.
To this end, participants were given precise instructions on attending to the paintings and text descriptions.
As shown in \autoref{fig:artwork_gallery}, two major patterns are induced by the walking directions \emph{LR} (left to right) and \emph{RL}  (right to left).
We further introduce subpatterns by varying the order between a painting and its text description. 
For more details, we refer to the dataset description \cite{darus-4023_2024}.
A summary of all five patterns is given in the following:\\[2ex]
\emph{Pattern 1:} Participants walked from \textbf{left to right} and attended each painting \textbf{before} its text description:

$F \rightarrow T_F \rightarrow R \rightarrow T_R \rightarrow H \rightarrow T_H \rightarrow B \rightarrow T_B \rightarrow C \rightarrow T_C.$\\[2ex]
\emph{Pattern 2:} Participants walked from \textbf{left to right} and attended each painting \textbf{after} its text description:

$T_F \rightarrow F \rightarrow T_R \rightarrow R \rightarrow T_H \rightarrow H \rightarrow T_B \rightarrow B \rightarrow T_C \rightarrow C.$\\[2ex]
\emph{Pattern 3:}
Participants walked from \textbf{right to left} and attended each painting \textbf{before} its text description:

$C \rightarrow T_C \rightarrow B \rightarrow T_B \rightarrow H \rightarrow T_H \rightarrow R \rightarrow T_R \rightarrow F \rightarrow T_F.$\\[2ex]
\emph{Pattern 4:}
Participants walked from \textbf{right to left} and attended each painting \textbf{after} its text description:

$T_C \rightarrow C \rightarrow T_B \rightarrow B \rightarrow T_H \rightarrow H \rightarrow T_R \rightarrow R \rightarrow T_F \rightarrow F.$\\[2ex]
\emph{Pattern 5:}
Participants attended paintings and text descriptions in random order; we only restricted them to attend each element exactly once.  

\subsection{Results}\label{subsec:results}
In this section, our NMF-based analysis approach is applied to the acquired mobile eye-tracking data, where two different scenarios are evaluated.
First, a random set of nine recordings is analyzed to showcase the usefulness of the visualization concepts. 
Subsequently, the whole dataset (27 recordings) is analyzed.
We used MATLAB (R2023b) to compute the results on 
a machine with AMD~Ryzen~9 3900X 12-core processor running at 3.79\,GHz and with~64\,GB~RAM.

\paragraph{Qualitative Evaluation}
This scenario shows the application of our approach to nine recordings. 
For the analysis, the stencil size of the gaze patches was $251\times 251$, the minimal fixation length $200$, and we used $k=8$ components.
As shown in \autoref{fig:teaser}, our approach successfully identifies four of the five AOIs, namely $R$, $B$, $H$, and $F$ ($C$ is not captured), with the five most dominant components. 
This can be observed in both the representative images and the shape and color of the spatial representations. 
The text descriptions $T_p$ are best represented by the last component but appear falsely in many other components. 
In general, this shows that our proposed sorting works quite well.

Analyzing only the first spatiotemporal component (\autoref{fig:vis_analysis_compts}) and considering the walking patterns (\autoref{fig:artwork_gallery}) regarding painting~$R$, one can identify the walking directions for each of the recordings, e.g., rec.~20, rec.~12, rec.~3, and rec.~22 seem to follow $LR$ (since the peak is rather at the beginning), and rec.~15, rec.~1, rec.~24, and rec.~14 follow $RL$ (since the peak is rather at the end of the recordings). 
Rec.~18 has a unique pattern as the peak is the starting point, indicating that the participant started with painting $R$.
To confirm these statements, the five AOIs were manually assigned as small sample images to the temporal indicator plots of the four most dominant components in~\autoref{fig:analysis_best9}. 
It can be observed that all statements are true (especially rec.~18 is correctly identified as the random walk -- Pattern 5) except for rec.~1.
Taking into account the representative image of rec.~1 (\autoref{fig:vis_analysis_compts}, purple frame), the highest peak refers to painting~$B$ instead of $R$, i.e., due to the color similarity between $B$ and $R$, the first area of high peaks refers to painting $R$.
Therefore, rec.~1 belongs to the other class of patterns, where the walking direction follows $LR$ (which is indeed~true).

The above visually guided analysis of the first component already provides important insights into the dataset.
The next steps of the analysis would be to successively consider other components of NMF and to combine the detected patterns of the individual components via additional visualizations.

\begin{figure*}[tbh]
\centering
\includegraphics[width=\textwidth]{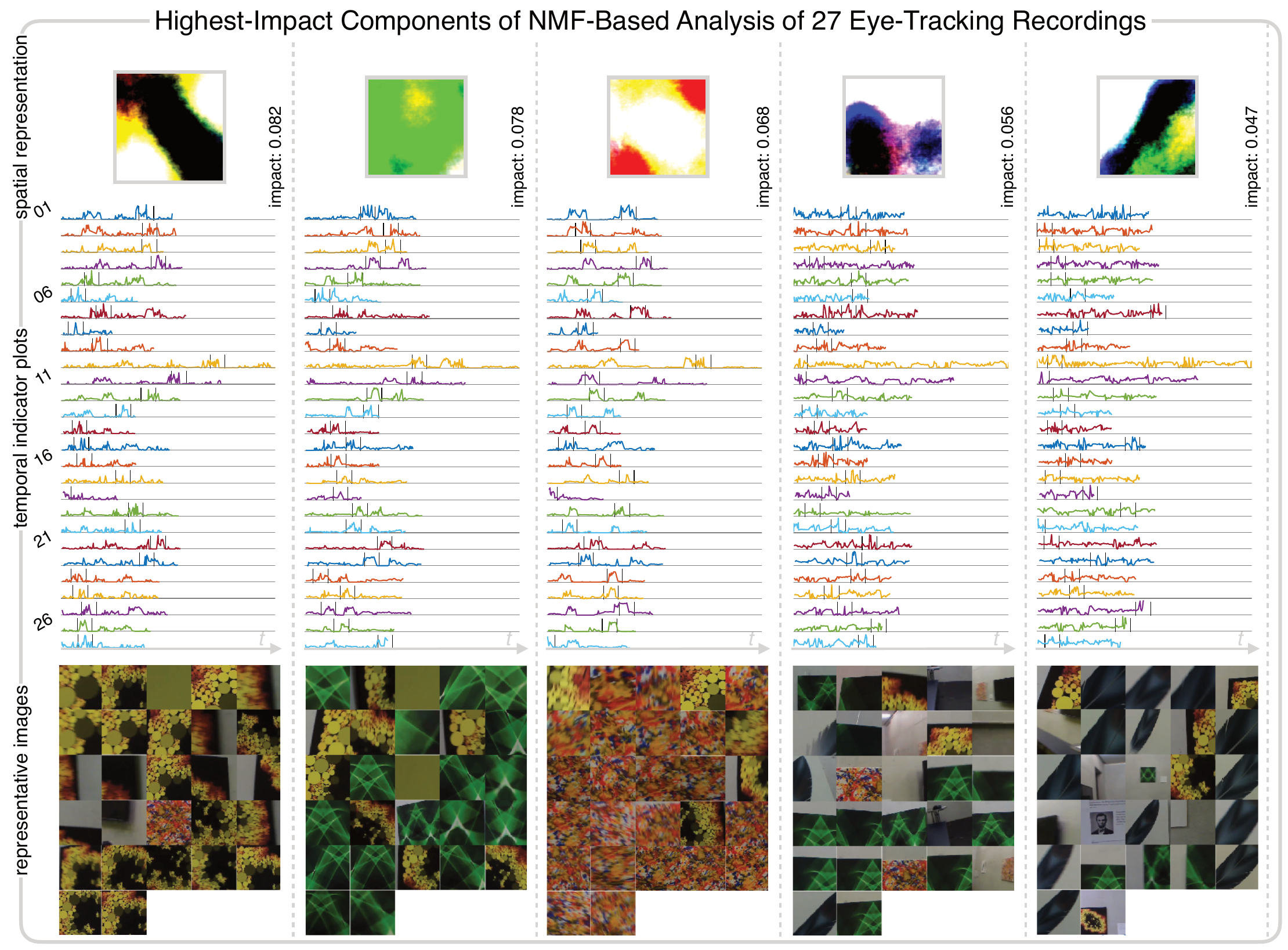}
\caption{Application of our NMF-based visual analysis to all 27 recordings of a scene from the art gallery. The preprocessed recordings are decomposed into ten spatiotemporal components, from which the five highest-impact components are shown. The spatial components and representative images characterize four of the five AOIs, and the temporal indicator plots show the temporal occurrences. The five lowest-impact components are illustrated in \autoref{fig:best27b} for reference. }
\label{fig:best27a}
\end{figure*}

\begin{figure*}[tbh]
\centering
\includegraphics[width=\textwidth]{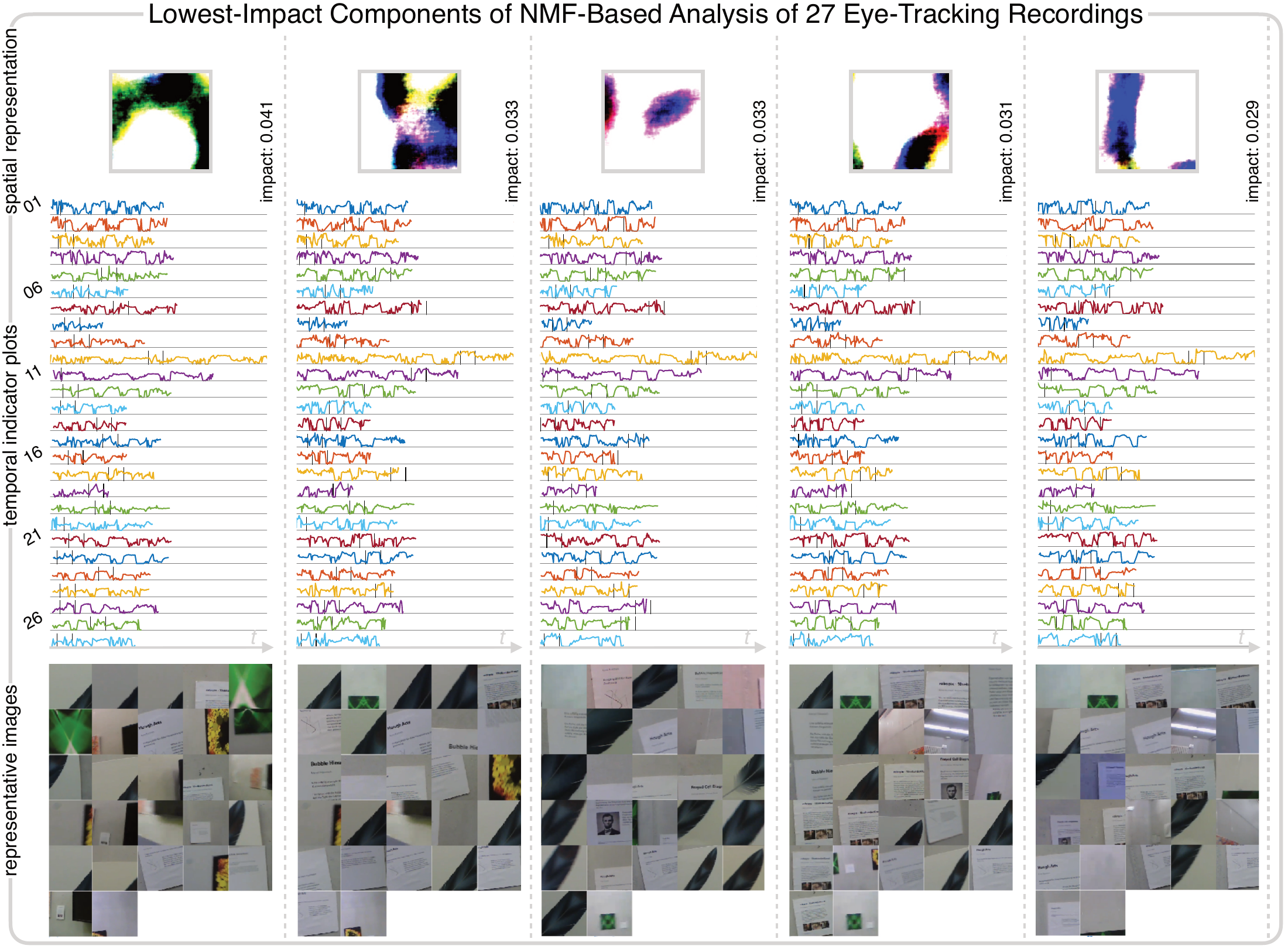}
\caption{NMF-based visual analysis of all 27 recordings of a scene from the art gallery. 
The preprocessed recordings are decomposed into ten spatiotemporal components, from which the five lowest-impact components are shown. For these components, the text descriptions are mostly depicted in the representative images. For comparison, we refer to the five highest-impact components, illustrated in \autoref{fig:best27a}.}
\label{fig:best27b}
\end{figure*}

\paragraph{Full Dataset Analysis}
The second scenario demonstrates the application of our approach to the full dataset (i.e., 27 recordings).
For the analysis, the stencil size of the gaze patches was $201\times 201$, the minimal fixation length $200$, and we used $k=10$ components. 
The visualization of the resulting $k=10$ components is divided into \autoref{fig:best27a} and \autoref{fig:best27b} for the five components with the highest and lowest impact, respectively.

Despite higher complexity, the analysis can be performed similarly to the previous paragraph (with nine recordings), leading to the same results.
It can be observed that the impact of components strongly decreases (from 0.082 to 0.029) and that important patterns occur in components 1--5 of \autoref{fig:best27a}.
Paintings $B$ and $R$ are characterized by components 1 and 3, and painting $H$ is identified by component~2.
Painting $F$ is mainly captured in component~4. 
A more thorough analysis of the individual temporal indicator plots reveals that clear peaks are visible and the assignment of walking directions (either $LR$ or $RL$) is possible.
For example, the three random walks (rec. 9, rec. 18, and rec. 27) can already be identified by inspecting component~3 because painting $R$ is clearly at the beginning or end (although placed centrally in~the~gallery).

As depicted via the decreasing impact of components, the components 6--10 in \autoref{fig:best27b} do not lead to an easy identification of the dataset patterns. This results from the similarities of shapes and colors of the text descriptions that are mostly identified in the representative images for the lowest-impact components.

\section{Discussion and Conclusion}\label{sec:conclusion}
Our preliminary investigations showed promising results for four of the five AOIs.
In the following, we discuss limitations regarding scalability~\cite{Richer2022scalability} and point out possible future work.

\paragraph{Scalability}
The \textit{algorithm scalability} depends on the combined duration of all recordings, the video resolution, and the chosen~$k$, which in turn depends on the number of AOIs to investigate, as our approach requires a higher $k$ to be able to represent more AOIs.
Through pre-processing, we reduce the amount of data to decrease data redundancy and runtime.
There exists an exact version of NMF that is NP-hard~\cite{Vavasis2010}.
The typical runtime for the MATLAB implementation we used in our experiments was within minutes on our preprocessed data (one frame per fixation).
Therefore, re-computing NMF cannot be done fast enough for acceptable interactivity in visual analysis, making it harder to compare different preprocessing parameterizations.

The \textit{visual scalability} depends on the same factors. 
With high $k$ (or many displayed components) and many recordings, images and line charts would get tiny when shown all at once, and interactive filtering/scrolling would be necessary.

\paragraph{Future Work}

We plan to extend our visual interface with interactions such as selecting a time (either in seconds or fixation index) to show corresponding video frames.
Visual scalability issues could be reduced through interaction or aggregation.
So far, we have only used case studies and a single dataset for evaluation; user studies with external eye-tracking researchers working with their own mobile recordings would be necessary for ecological validity.

\begin{acks}
This work was supported by the Deutsche Forschungsgemeinschaft (DFG, German Research Foundation) under
Project-ID 270852890 -- GRK 2160/2, 
Project-ID 251654672 -- TRR 161, 
Project-ID 279064222 -- SFB 1244, 
Project-ID 390831618 -- EXC 2120/1, 
and Project-ID 449742818,
as well as by the Cyber Valley Research Fund. 

Furthermore, we would like to express our gratitude to the organizers of the SFB-TRR 161 hackathon ``Dimensionality Reduction for Eye-Tracking,'' in which the initial idea for this project evolved.
\end{acks}

\bibliographystyle{ACM-Reference-Format}
\bibliography{main}

\end{document}

%% file: own_packages_and_definitions.tex
\usepackage{enumitem}
\usepackage{subcaption}